\documentclass[conference]{IEEEtran}
\IEEEoverridecommandlockouts

\usepackage{cite}
\usepackage{amsmath,amssymb,amsfonts}
\usepackage{graphicx}
\usepackage{textcomp}
\usepackage{xcolor}
\usepackage{booktabs}
\usepackage{multirow}
\usepackage{enumitem}
\usepackage[hidelinks]{hyperref}
\usepackage{cleveref}

\def\BibTeX{{\rm B\kern-.05em{\sc i\kern-.025em b}\kern-.08em
    T\kern-.1667em\lower.7ex\hbox{E}\kern-.125emX}}

\providecommand{\Description}[1]{}
\renewcommand{\footnoterule}{\kern -3pt \hrule width 0.4\columnwidth \kern 2.6pt}
\makeatletter
\long\def\@makefntext#1{\noindent #1}
\makeatother

\begin{document}

\title{\vspace{-2pt}WorldMAP: Bootstrapping Vision-Language Navigation Trajectory Prediction with Generative World Models}

\author{
\IEEEauthorblockN{
Hongjin Chen$^{1,5,*}$,
Shangyun Jiang$^{2,*}$,
Tonghua Su$^{1,\dagger}$,
Chen Gao$^{3,5,\dagger}$,
Xinlei Chen$^{3}$,
Yong Li$^{3}$,
Zhibo Chen$^{4,5,\dagger}$\\[3pt]
}
\IEEEauthorblockA{
$^{1}$Harbin Institute of Technology \quad
$^{2}$Shandong University \quad
$^{3}$Tsinghua University \\
$^{4}$University of Science and Technology of China \quad
$^{5}$Zhongguancun Academy \\
}
\thanks{$^{*}$Authors contributed equally to this research. Shangyun Jiang conducted this work during an internship at Tsinghua University.\newline $^{\dagger}$Corresponding authors: Tonghua Su (thsu@hit.edu.cn), Chen Gao (chgao96@gmail.com), and Zhibo Chen (chenzhibo@ustc.edu.cn); Chen Gao and Zhibo Chen are the project leaders.}
}

\pagestyle{plain}
\maketitle
\thispagestyle{plain}

\begin{abstract}
Vision-language models (VLMs) and generative world models are opening new opportunities for embodied navigation. VLMs are increasingly used as direct planners or trajectory predictors, while world models support look-ahead reasoning by imagining future views. Yet predicting a reliable trajectory from a single egocentric observation remains challenging. Current VLMs often generate unstable trajectories, and world models, though able to synthesize plausible futures, do not directly provide the grounded signals needed for navigation learning. This raises a central question: how can generated futures be turned into supervision for grounded trajectory prediction? We present WorldMAP, a teacher--student framework that converts world-model-generated futures into persistent semantic-spatial structure and planning-derived supervision. Its world-model-driven teacher builds semantic-spatial memory from generated videos, grounds task-relevant targets and obstacles, and produces trajectory pseudo-labels through explicit planning. A lightweight student with a multi-hypothesis trajectory head is then trained to predict navigation trajectories directly from vision-language inputs. On Target-Bench, WorldMAP achieves the best ADE and FDE among compared methods, reducing ADE by 18.0\% and FDE by 42.1\% relative to the best competing baseline, while lifting a small open-source VLM to DTW performance competitive with proprietary models. More broadly, the results suggest that, in embodied navigation, the value of world models may lie less in supplying action-ready imagined evidence than in synthesizing structured supervision for navigation learning.\end{abstract}


\section{Introduction}
\label{sec:intro}

\begin{figure}[t]
  \centering
  \includegraphics[width=0.95\columnwidth]{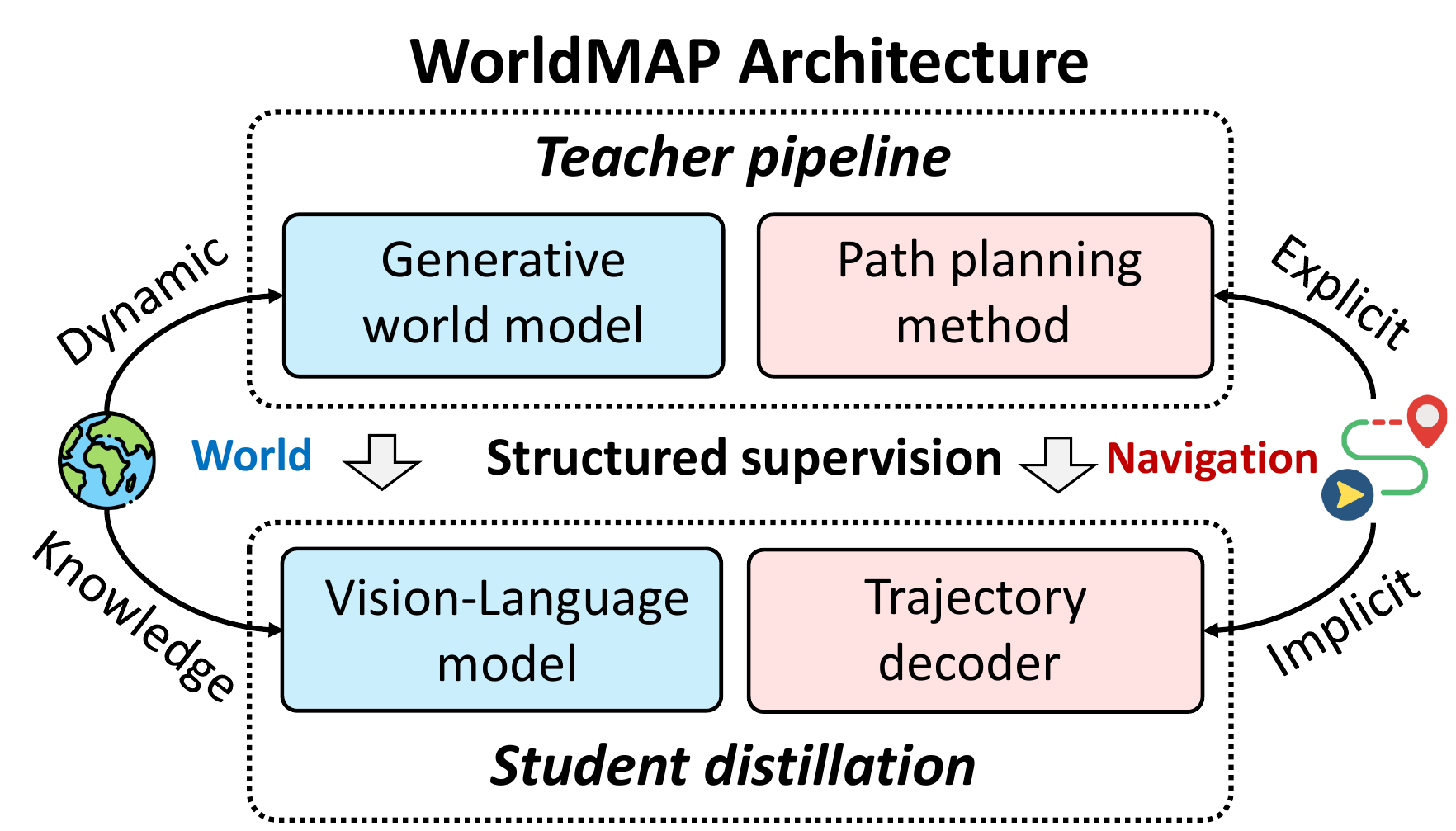}
  \caption{\textbf{Teacher--student distillation in WorldMAP.} A world-model-driven teacher converts generated futures into grounded training signals for a lightweight vision-language student, which learns to predict navigation trajectories directly from the observation and instruction.}
  \Description{A conceptual teacher-student distillation diagram for WorldMAP. The top row shows a teacher pipeline with a generative world model and an explicit planner producing grounded training signals. The bottom row shows student distillation with a vision-language model and a trajectory decoder. The figure contrasts explicit teacher-side planning with an efficient learned student predictor.}
  \label{fig:distill}
\end{figure}

\begin{figure*}[t]
  \centering
  \includegraphics[width=0.92\textwidth,trim=8 0 8 6,clip]{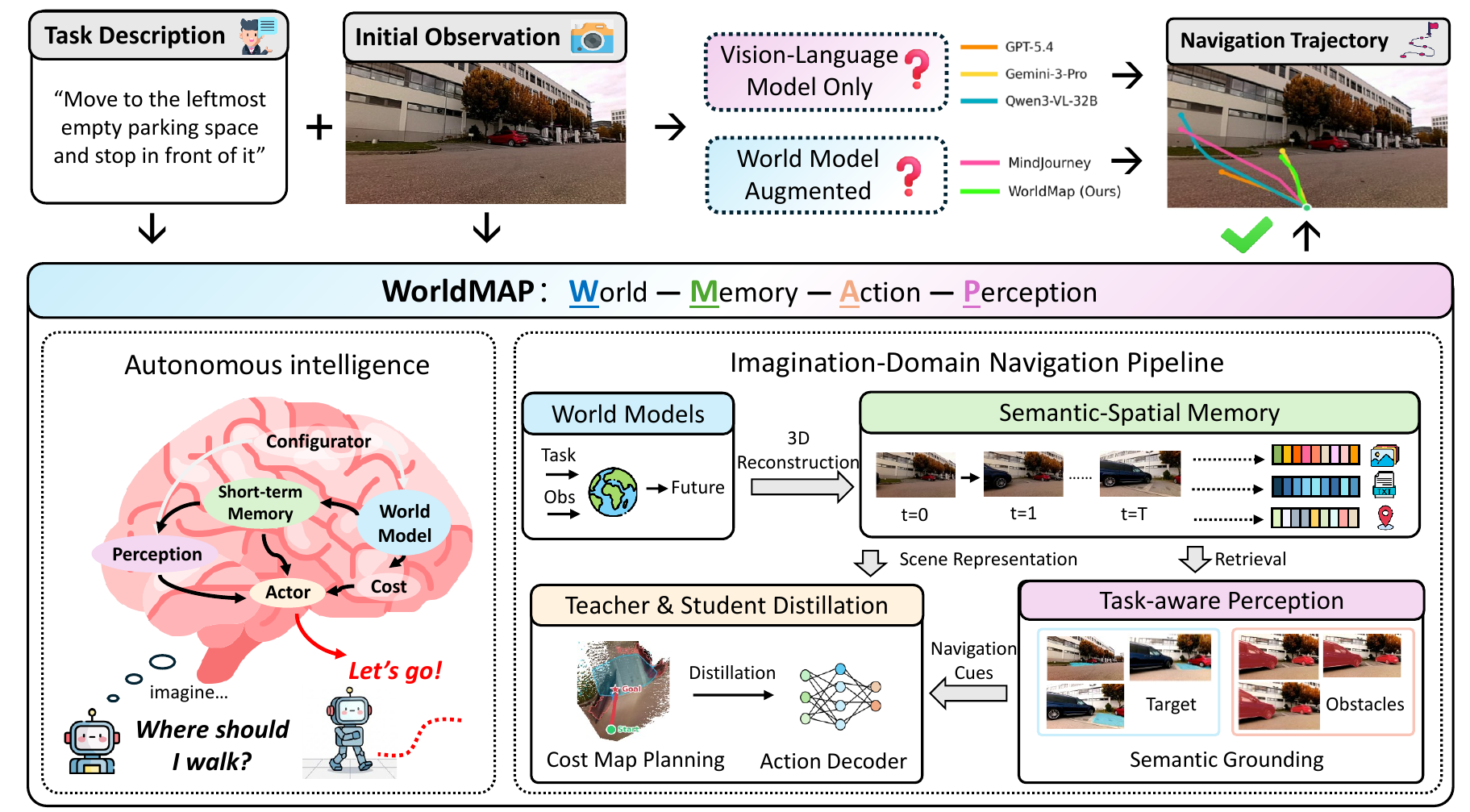}
  \caption{\textbf{Motivation and design rationale of WorldMAP.} \textit{Top}: VLM-only prediction and world-model-augmented reasoning remain unreliable for single-observation trajectory prediction. \textit{Bottom}: WorldMAP organizes navigation around a \textit{World-Memory-Action-Perception} decomposition, inspired by LeCun's architecture for autonomous machine intelligence~\cite{lecun2022path}, and converts generated futures into persistent semantic-spatial structure and planning-derived supervision for a lightweight student predictor.}
  \Description{A two-part overview figure for WorldMAP. The top row shows a task description, a single initial observation, baseline predictions from VLM-only and world-model-augmented approaches, and the improved navigation trajectory produced by WorldMAP. The bottom row illustrates the World-Memory-Action-Perception decomposition and the corresponding navigation pipeline, including world models, semantic-spatial memory, task-aware perception, planning, and student distillation.}
  \label{fig:storytelling}
\end{figure*}

Reliable navigation from egocentric observations and natural-language instructions remains a central problem in embodied AI~\cite{duan2022survey, liu2025aligning}, especially in continuous, previously unseen environments. In such settings, an agent must ground language in the scene, infer traversable space from limited visual evidence, and generate a physically plausible path toward a semantically specified target, rather than choose among a small set of pre-defined viewpoints~\cite{anderson2018vision, krantz2020beyond, krantz2021waypoint, hong2022bridging}. This makes trajectory prediction fundamentally more demanding than discrete action selection in conventional vision-language navigation settings~\cite{krantz2021waypoint, hong2022bridging}.

Classical navigation pipelines address this challenge through explicit localization, mapping, and planning, often with SLAM-style geometric maps, semantic maps, and analytical path planners~\cite{cadena2017past, mur2015orb, chaplot2020learning, chaplot2020object}. These systems remain effective because they expose interpretable intermediate representations that support reliable motion planning and obstacle avoidance. However, they often depend on persistent map construction, repeated scene exploration, or carefully engineered modular designs, which can be cumbersome in previously unseen or dynamically changing environments. At the same time, language-grounded navigation research suggests that explicit spatial representations remain valuable even in learned embodied agents, including top-down semantic maps and structured 3D environment representations~\cite{georgakis2022cross, liu2024volumetric}.

Recent progress has opened two complementary alternatives to fully explicit map-building pipelines. One line studies navigation with large vision-language models (VLMs), using them as high-level planners, action selectors, or end-to-end embodied policies~\cite{nasiriany2024pivot, goetting2025end, nie2025wmnav}. Another explores world models that predict future observations or latent scene evolution to support look-ahead planning, imagined rollouts, or mental simulation~\cite{koh2021pathdreamer, wang2023dreamwalker, bar2025navigation, yao2025navmorph}. Related navigation-specific work further asks whether predicted future-view semantics or generated visual imaginations can help downstream decisions~\cite{li2023improving, perincherry2025visual}. Together, these advances suggest that semantic priors and imagined future views can partially compensate for incomplete direct observation.

Yet reliable navigation trajectory prediction remains unresolved. Recent benchmark evidence suggests that current VLMs still struggle to produce stable and grounded navigation traces from a single egocentric observation, especially when the output must remain traversable, respect scene geometry, and stop at the correct semantic target~\cite{windecker2025navitrace}. Meanwhile, world-model-generated viewpoints can improve some spatial reasoning tasks when used as additional multi-view evidence, but their benefit is not uniform. Recent studies show that imagination can be helpful in some settings~\cite{yang2025mindjourney, cao2025spatialdreamer}, while excessive or ungated imagination may also introduce misleading evidence and degrade downstream performance~\cite{yu2026and}.
This reveals a key gap. VLMs are strong at language understanding and semantic grounding, but they remain unstable as direct trajectory predictors from raw egocentric input. World models can synthesize visually plausible future views, yet those views do not by themselves provide the structured supervision needed for grounded trajectory learning. What is missing is therefore not merely additional imagined evidence, but a mechanism that converts generated futures into persistent semantic and geometric structure that can serve as structured supervision for learnable, grounded trajectory prediction.
To address this gap, we present \textbf{WorldMAP}, a teacher--student framework for navigation trajectory prediction inspired by knowledge distillation~\cite{hinton2015distilling}, as summarized in Figure~\ref{fig:distill}. Its core idea is to use a world-model-driven teacher to convert generated future videos into semantic-spatial memory, task-aware grounding, and explicit planning signals, and to distill the resulting planning-derived trajectory pseudo-labels into a lightweight student predictor. Given a single observation and a language instruction, the teacher identifies target regions and navigation-relevant obstacles across generated views, projects them into a shared navigation plane, and derives a grounded trajectory through explicit planning. The student then predicts navigation trajectories directly from vision-language inputs, without re-running the teacher pipeline at test time. In this way, WorldMAP uses world models not as direct action policies, but as supervision engines for grounded navigation learning. As illustrated in Figure~\ref{fig:storytelling}, current VLM-only prediction and world-model-augmented reasoning remain insufficiently reliable for single-observation trajectory prediction, motivating the \textit{World-Memory-Action-Perception} decomposition underlying WorldMAP.
More broadly, WorldMAP repositions the role of world models in embodied navigation. To our knowledge, it is the first work to explicitly use world-model generation as supervision rather than test-time evidence for single-observation grounded navigation trajectory prediction.
The main contributions of this work are summarized as follows.
\begin{itemize}
    \item \textbf{A New Teacher-Student Distillation Framework for Navigation Trajectory Prediction.} We introduce a teacher--student framework that distills planning-derived trajectory supervision into a lightweight predictor operating directly on vision-language inputs.
    \item \textbf{A World-Model-Driven Engine for Grounded Trajectory Generation.} We develop a pipeline that converts generated futures into semantic-spatial memory, task-aware target and obstacle grounding, and explicit geometric planning signals, turning world-model outputs into learnable supervision rather than downstream evidence.
    \item \textbf{Strong Performance on Target-Bench.} On Target-Bench, a benchmark for navigation toward semantic targets in unstructured real-world environments~\cite{wang2025target}, WorldMAP achieves the best ADE and FDE among the compared methods while enabling a small open-source VLM backbone to perform competitively with much stronger proprietary models.
  \end{itemize}

\section{Related Work}

\subsection{Structured Spatial Representations for Navigation}

Classical robot navigation has long relied on explicit localization, mapping, and planning over geometric spatial representations. Embodied navigation extends this line with learned semantic maps and structured 3D environment representations for language grounding and action planning\cite{thrun2002probabilistic,cadena2017past,mur2015orb,chaplot2020learning,chaplot2020object,georgakis2022cross,liu2024volumetric}. 
Such structured representations provide interpretable intermediate state, reliable geometric control, and strong support for traversability-aware planning. 
However, they often depend on persistent map construction, repeated scene interaction, or online state estimation, which is less natural in single-observation navigation settings where an agent must infer a plausible path from limited egocentric evidence.

\subsection{World Models and Vision-Language Reasoning for Navigation}

\noindent\textbf{World models for navigation.}
World models have been increasingly explored as predictive engines for embodied navigation, where future observations or latent scene dynamics are imagined to support planning beyond the current view \cite{koh2021pathdreamer,wang2023dreamwalker,bar2025navigation,yao2025navmorph}. 
PathDreamer synthesizes unobserved panoramic observations for downstream route evaluation \cite{koh2021pathdreamer}, while DREAMWALKER performs mental planning in an abstract world model for continuous vision-language navigation \cite{wang2023dreamwalker}. 
More recent work studies controllable future prediction and adaptive world modeling for navigation, including Navigation World Models \cite{bar2025navigation} and NavMorph \cite{yao2025navmorph}. 
At the same time, Target-Bench shows that mapless path planning toward semantic targets from a single observation remains challenging for current world models, even when generated views appear plausible \cite{wang2025target}.

\noindent\textbf{Vision-language models for navigation.}
Recent work has also explored large vision-language models as embodied decision-makers for navigation. 
LM-Nav composes pre-trained language, vision, and navigation modules for language-conditioned robotic navigation \cite{shah2023lm}, VLMnav treats navigation as end-to-end action selection with a VLM \cite{goetting2025end}, and PIVOT uses iterative visual prompting to let a VLM choose among progressively refined action, localization, or trajectory proposals \cite{nasiriany2024pivot}. 
Related hybrid designs further combine language reasoning with predictive environment modeling; for example, WMNav integrates VLM reasoning into a world-model-based loop for object-goal navigation \cite{nie2025wmnav}. 
However, benchmark evidence from NaviTrace suggests that even strong contemporary VLMs still exhibit substantial errors in spatial grounding and goal localization when asked to output navigation traces directly in image space \cite{windecker2025navitrace}.

\noindent\textbf{Test-time reasoning with generated views.}
A closely related line augments vision-language reasoning with generated or imagined viewpoints at test time. 
ImagineNav uses imagined future views to turn mapless navigation into a viewpoint-selection problem for VLM-based decision making \cite{zhaoimaginenav}. 
MindJourney couples a VLM with a controllable world model to gather multi-view imagined evidence for spatial reasoning \cite{yang2025mindjourney}, while AVIC shows that the benefit of imagination depends on when it is invoked and how much imagined evidence is used, rather than increasing world-model calls indiscriminately \cite{yu2026and}. 
Compared with prior work, which mainly uses generated futures as transient evidence for planning or test-time reasoning, our method repositions world-model generation as a source of persistent semantic-spatial structure and, more importantly, as a mechanism for synthesizing structured supervision for grounded navigation learning.

\section{Method}
\label{sec:method}

\subsection{Overview}
\label{sec:method-overview}

\begin{figure*}[t!]
  \centering
  \includegraphics[width=\textwidth]{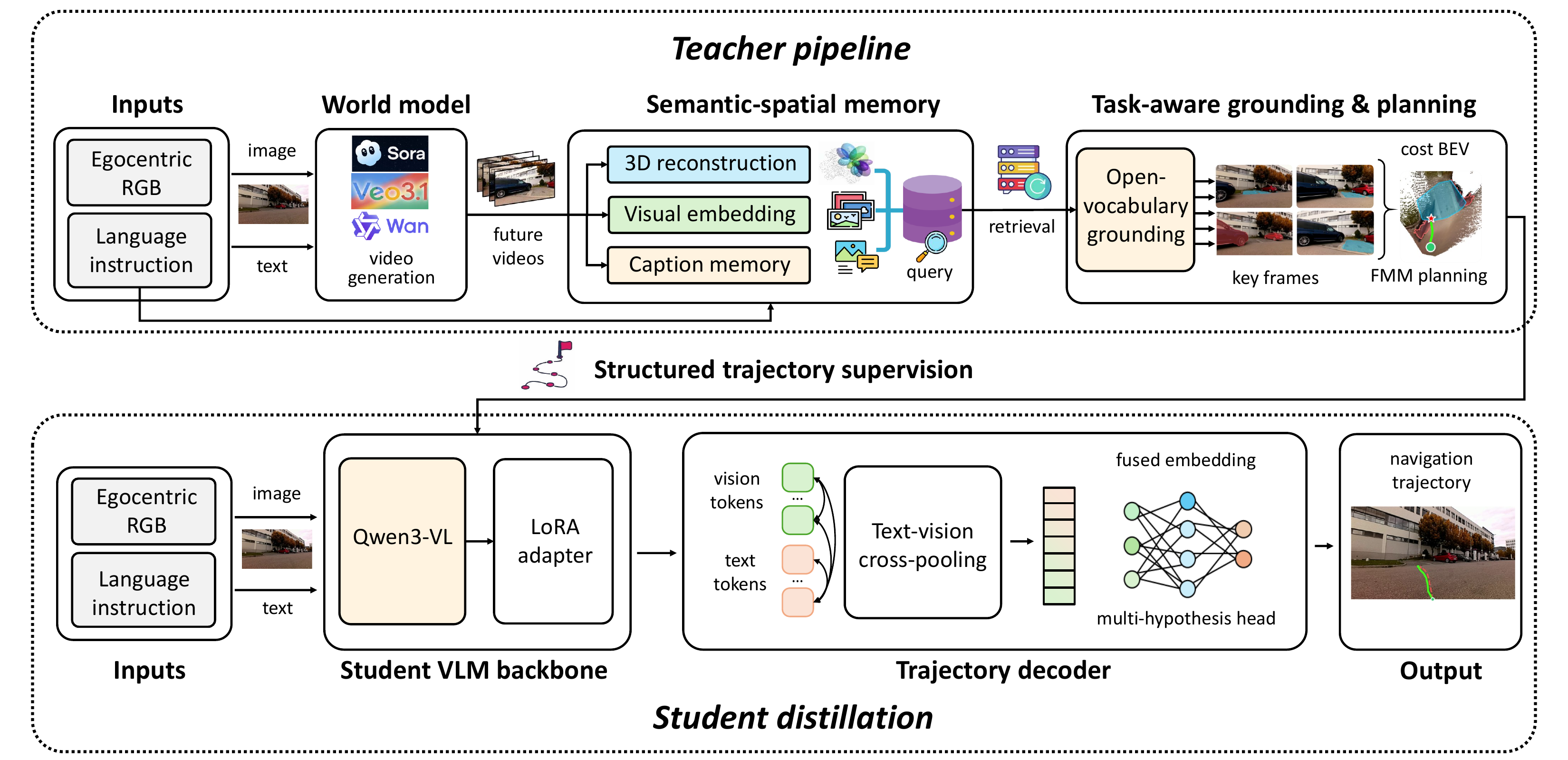}
  \caption{\textbf{Architecture of WorldMAP.} A world-model-driven teacher converts generated futures into semantic-spatial memory, task-aware grounding, and cost-BEV planning, and then produces trajectory pseudo-labels for student training. At inference time, only the lightweight student is used to predict the final navigation trajectory.}
  \Description{A teacher-student architecture diagram for WorldMAP. The teacher takes a first-person RGB image and a language instruction, uses world-model-generated future videos, consolidates them into semantic-spatial memory with 3D reconstruction, visual embeddings, and caption memory, then performs task-aware grounding and FMM planning on a cost BEV to generate trajectory pseudo-labels. The student takes vision and language inputs, processes them with a VLM backbone and a trajectory decoder with cross-pooling and a multi-hypothesis head, and outputs a navigation trajectory.}
  \label{fig:framework}
\end{figure*}

Following the \textit{World-Memory-Action-Perception} decomposition introduced in \cref{sec:intro}, WorldMAP consists of two coupled components: a \emph{world-model-driven teacher} and a lightweight \emph{student trajectory predictor} (\cref{fig:framework}).

Given a first-person RGB observation and a natural-language instruction, the teacher first synthesizes a short future video and consolidates it into semantic-spatial memory with associated visual embeddings and captions. This memory is then queried to ground the target and identify navigation-relevant obstacles. The grounded multi-view evidence is projected onto a shared navigation plane, converted into a cost map, and used by an explicit planner to generate structured trajectory supervision. The student is finally trained to predict navigation trajectories directly from vision-language inputs using this supervision.

The central design choice is to separate \emph{semantic grounding} from \emph{geometric planning}. Semantics are extracted from the generated multi-view video through memory retrieval and VLM-guided perception, while geometry is expressed in a common navigation plane and BEV coordinate system. This design allows the teacher to convert generated futures into structured supervision rather than forcing either the world model or the VLM to predict trajectories directly from raw egocentric evidence.
\subsection{World-Model-Driven Trajectory Teacher}
\label{sec:method-teacher}

The teacher converts generated future observations into structured trajectory supervision through three stages: world construction, memory-guided perception, and explicit geometric planning.

\subsubsection{World Construction}
\label{sec:method-world}

The teacher begins by converting a single observation into two complementary outputs: a 3D reconstruction for planning and a semantic-spatial memory for retrieval.

\noindent\textbf{Future video generation.}
We leverage multiple streams of world-model-generated future videos. Although individual generations can be imperfect, they offer complementary nearby viewpoints and scene hypotheses that are useful for downstream grounding and map construction.

\noindent\textbf{3D reconstruction and memory construction.}
For each generated frame, we estimate monocular depth using Depth Anything 3~\cite{lin2025depth} and backproject it into a shared scene representation. In parallel, each frame is encoded into semantic-spatial memory with its visual embedding, caption, and camera pose. Together, these two representations provide the geometric and semantic support needed for later grounding and planning.

\noindent\textbf{Navigation plane.}
To keep downstream geometry consistent, we estimate a scene-level ground plane and use it as the shared navigation plane for projection, BEV construction, and trajectory generation. This plane serves as the common spatial reference for both teacher planning and student supervision.
\subsubsection{Task-Aware Grounding}
\label{sec:method-grounding}

Given the task instruction and the semantic-spatial memory, the teacher next determines \emph{where the target is} and \emph{which scene elements should be avoided}. This grounding step is performed before planning so that the planner operates on explicit entities rather than raw image evidence.

\noindent\textbf{Target grounding.}
\label{sec:method-goal}
We first retrieve the top-ranked memory frames that are most relevant to the instruction using CLIP-based visual-text similarity~\cite{radford2021learning}. These candidate frames provide multiple viewpoints of the likely target region. A VLM then summarizes the task into a target description conditioned on the retrieved frames and their memory context, which is passed to the open-vocabulary segmentation model UniPixel~\cite{liu2025unipixel} to predict target masks. The selected masks are fused in 3D and projected into the BEV coordinate system, producing a grounded target region rather than a single image-space point.

\noindent\textbf{Obstacle grounding.}
We use the same memory to identify navigation-relevant obstacles. A VLM reads the task together with retrieved frame captions and proposes object categories that may physically block the path to the target. These categories are then used to retrieve additional supporting views and drive open-vocabulary segmentation with UniPixel~\cite{liu2025unipixel}. The resulting obstacle masks are projected into the same BEV coordinate system as the target region.

Target grounding is task-specific and tied to the instruction semantics, whereas obstacle grounding is path-specific and emphasizes objects that interfere with traversal. Both are ultimately expressed in the same geometric frame for downstream planning.
\subsubsection{Geometric Planning and Supervision}
\label{sec:method-bev-render}

We next convert grounded multi-view evidence into a top-down planning representation for teacher trajectory generation.

\subsubsection{Cost BEV Construction}
To unify grounding and planning, we project grounded multi-view evidence onto a shared navigation plane and accumulate it in a top-down BEV grid. For each generated frame, we backproject depth pixels into world coordinates and accumulate their RGB values on this grid. Only points within a ground-level height band are retained, which keeps traversable surfaces and low obstacles while suppressing high structures and sky artifacts. When multiple frames observe the same BEV cell, their colors are fused using confidence-aware averaging.

The result is a metric-preserving RGB BEV that aggregates coverage from multiple future views. We then overlay the grounded target and obstacle regions so that target, obstacles, and free space are all expressed in the same coordinate system.

\subsubsection{Cost-Aware Planning and Supervision}
\label{sec:method-cost-plan}
\noindent\textbf{Cost map construction.}
We convert this BEV representation into a planning cost map. Cells corresponding to explicit obstacle masks are blocked. In addition, unobserved background regions are treated conservatively as non-traversable structure, since they typically correspond to walls or areas unsupported by the generated views. Around blocked cells, we apply a safety margin and assign higher traversal costs to near-obstacle regions, encouraging paths through the middle of free corridors.

\noindent\textbf{FMM planning.}
Given the start position and the grounded target region, we run the Fast Marching Method (FMM)~\cite{sethian1996fast, sethian1999fast} on the BEV cost map to obtain a minimum-cost path from start to goal. The raw grid path is then smoothed and resampled into a sparse waypoint sequence.

These BEV waypoints are finally mapped back onto the navigation plane to form the teacher trajectory. In this way, explicit planning serves not as the final navigation policy, but as the mechanism that converts grounded scene structure into structured trajectory supervision.
\subsection{Student Trajectory Learning}
\label{sec:method-student}

The student is a lightweight vision-language trajectory predictor trained on the supervision generated by the teacher. At inference time, it takes a first-person observation and a language instruction as input and predicts the final navigation trajectory directly, without re-running the teacher pipeline.

\noindent\textbf{Architecture.}
The student uses a compact trajectory decoder on top of multimodal conditioning features. Text tokens, vision tokens, and their pooled cross-modal summary are fused into a compact representation, which is then decoded into future waypoints. To better absorb the ambiguity present in teacher supervision, the decoder predicts $K$ trajectory hypotheses rather than a single trajectory, together with a confidence score for each hypothesis.

\noindent\textbf{Training supervision.}
The student is trained from trajectories generated by the world-model-driven teacher. This supervision is richer than direct image-to-trajectory regression because it already encodes target grounding, obstacle awareness, and explicit geometric planning. The role of the student is therefore not to rediscover these structures from scratch, but to distill them into a compact predictor that runs from vision-language inputs alone.

\begin{figure}[t]
  \centering
  \includegraphics[width=\columnwidth]{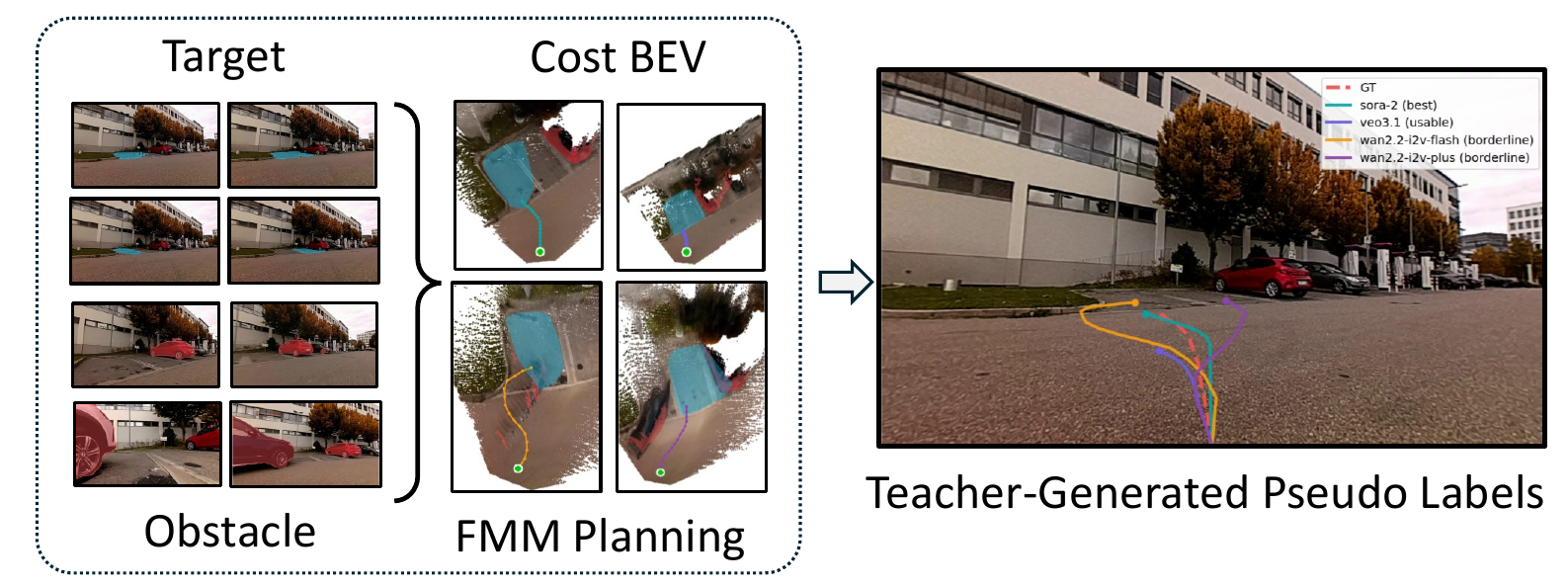}
  \caption{\textbf{Teacher-generated trajectory pseudo-labels from world-model futures.} The teacher grounds targets and obstacles across generated views, projects them into a shared BEV planning space, constructs a cost map, and runs FMM to obtain trajectory pseudo-labels for student training.}
  \Description{An illustration of teacher-generated trajectory pseudo-labels in WorldMAP. The left panel shows target and obstacle grounding across multiple generated views. The middle panel shows the BEV planning space, cost map construction, and FMM path planning. The right panel shows projected pseudo-label trajectories overlaid on the original image together with the ground-truth trajectory.}
  \label{fig:wm-pseudo-labels}
\end{figure}

\noindent\textbf{Training objective.}
Let $\hat{\mathbf{Y}}=\{\hat{\mathbf{Y}}^{(k)}\}_{k=1}^{K}$ denote the $K$ predicted waypoint sequences and let $\mathbf{Y}=\{\mathbf{y}_t\}_{t=1}^{T}$ be the target trajectory. For each hypothesis, WorldMAP combines a waypoint regression term with a segment-direction consistency term:
\begingroup
\setlength{\abovedisplayskip}{4pt}
\setlength{\abovedisplayshortskip}{4pt}
\small
\begin{equation}
\label{eq:student_loss}
\mathcal{L}^{(k)} = \frac{1}{T}\sum_{t=1}^{T}\lVert \hat{\mathbf{y}}^{(k)}_t-\mathbf{y}_t \rVert_2 + \frac{\lambda_d}{T}\sum_{t=1}^{T}\Bigl(1-\cos(\Delta\hat{\mathbf{y}}^{(k)}_t,\Delta\mathbf{y}_t)\Bigr).
\end{equation}
\endgroup
The first term is a per-waypoint $\ell_2$ regression loss, while the second term encourages local segment-direction consistency, where $\Delta\mathbf{y}_t=\mathbf{y}_t-\mathbf{y}_{t-1}$ denotes the trajectory segment at step $t$.
The final supervision uses the best-matching hypothesis,
\begin{equation}
\label{eq:best_of_k}
\mathcal{L}_{\mathrm{student}} =
\frac{1}{B}\sum_{i=1}^{B}\min_{k \in \{1,\dots,K\}} \mathcal{L}^{(k)}_i,
\end{equation}
where $\lambda_d=0.5$ in our implementation. This objective encourages one hypothesis to align closely with the target trajectory while preserving flexibility during learning, and the direction term helps maintain local trajectory shape beyond pointwise regression alone. During evaluation, a single trajectory is selected from the predicted hypotheses as the final output.

\section{Experiments}
\label{sec:experiments}

\subsection{Experimental Setup}

\noindent\textbf{Benchmark.}
We evaluate on Target-Bench~\cite{wang2025target}, a real-world benchmark for evaluating world models on path planning toward semantic targets in unstructured indoor and outdoor environments. Each sample provides a first-person RGB observation, a natural-language navigation instruction, and a SLAM-estimated 3D trajectory of a quadruped robot. Following our preprocessing pipeline, we project the robot trajectory onto the corresponding real first-frame image to obtain a 2D pixel-space trajectory, which is used as ground truth for final evaluation.

\noindent\textbf{Metrics.}
We report trajectory error in 2D pixel space. Let $\hat{\mathbf{P}}=\{\hat{\mathbf{p}}_t\}_{t=1}^{T}$ and $\mathbf{P}=\{\mathbf{p}_t\}_{t=1}^{T}$ denote the predicted and ground-truth trajectories projected onto the real first-frame image. Average Displacement Error (ADE) is defined as
\begin{equation}
\mathrm{ADE}=\frac{1}{T}\sum_{t=1}^{T}\left\lVert \hat{\mathbf{p}}_t-\mathbf{p}_t \right\rVert_2,
\end{equation}
Final Displacement Error (FDE) is defined as
\begin{equation}
\mathrm{FDE}=\left\lVert \hat{\mathbf{p}}_T-\mathbf{p}_T \right\rVert_2,
\end{equation}
and normalized Dynamic Time Warping (DTW) is defined as
\begin{equation}
\mathrm{DTW}_{\mathrm{norm}}=\frac{1}{L}\,\mathrm{DTW}(\hat{\mathbf{P}},\mathbf{P}),
\end{equation}
where $L$ denotes the DTW warping-path length. All three metrics are computed on the projected 2D trajectories in the real first-frame image, and lower values indicate better performance.

\noindent\textbf{Implementation.}
Unless otherwise noted, the reported main results correspond to the final WorldMAP student. The teacher leverages multiple streams of world-model-generated future videos to generate structured trajectory supervision through semantic-spatial grounding and explicit planning. The student is a lightweight predictor built on a small open-source VLM backbone and trained from teacher pseudo-labels together with benchmark GT when available. The ablations below vary only the composition of this supervision while keeping the student architecture fixed.
\subsection{Main Results}
\label{sec:exp-main}
WorldMAP is evaluated as a trained student predictor, whereas proprietary/open-source VLM baselines and MindJourney are evaluated as direct or test-time reasoning systems under the same benchmark protocol.
We compare three families of methods: proprietary VLMs (GPT-5.4, o3, and Gemini-3-Pro), open-source VLMs used as direct trajectory predictors (Qwen3-VL-4B/8B/32B and InternVL3-14B), and world-model-augmented methods, including MindJourney and WorldMAP. This comparison addresses two related questions: whether direct VLM trajectory prediction is already sufficient in this benchmark, and whether world models are more effective when used to generate training signals rather than merely to provide additional imagined evidence for downstream reasoning.

\noindent\textbf{Evaluation in projected 2D image space (\cref{tab:main-2d}).}

WorldMAP achieves the best ADE and FDE among all compared methods. Relative to the best competing baseline, Gemini-3-Pro, it reduces ADE from 51.27 to 42.06 (18.0\%) and FDE from 67.19 to 38.87 (42.1\%), while remaining close on normalized DTW (31.95 vs.\ 31.63). Taken together, these results reveal a clear pattern: direct trajectory prediction remains unreliable even for strong VLMs, whereas training a small open-source backbone with teacher-generated, planning-derived pseudo-labels lift a small open-source backbone into a much stronger performance regime.
Notably, the o3-based MindJourney model also underperforms the direct o3 baseline on all three metrics, indicating that additional imagined views are not automatically beneficial in this benchmark.

\begin{figure*}[!tbp]
  \centering
  \includegraphics[width=0.94\textwidth]{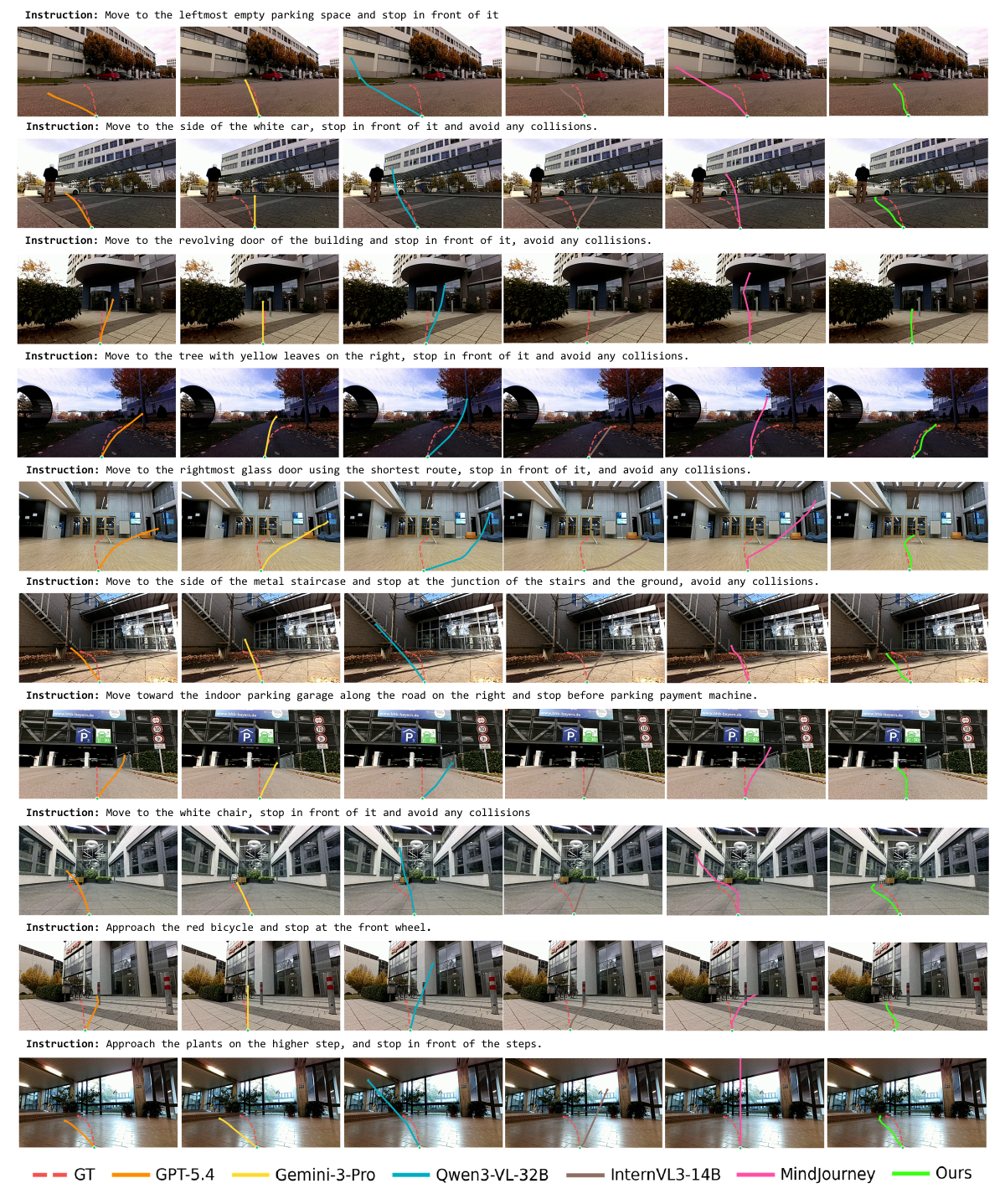}
  \caption{\textbf{Qualitative comparison on Target-Bench.} Each row shows one instruction and the projected trajectories from the ground truth and competing methods in the real first-frame image. WorldMAP more consistently follows traversable floor geometry and stops closer to the intended target, whereas baselines are more prone to drift, overshoot, or pass through non-traversable regions.}
  \Description{A qualitative comparison figure for navigation tasks. Each row shows an instruction and projected trajectories from the ground truth and competing methods in the real first-frame image, illustrating that WorldMAP produces more grounded trajectories than direct VLM and world-model-augmented baselines.}
  \label{fig:qualitative}
\end{figure*}

Figure~\ref{fig:qualitative} shows that these gains are not only numerical. Compared with direct-prediction and world-model-augmented baselines, WorldMAP more consistently follows traversable floor geometry, avoids implausible shortcuts, and stops closer to the intended target.

\begin{table}[t]
  \centering
  \caption{\textbf{Main results in projected 2D image space on Target-Bench.}\\
  \parbox[t]{\columnwidth}{\normalfont\raggedright ADE, FDE, and normalized DTW are computed on projected trajectories in the real first-frame image. Lower is better for all metrics. MindJourney$^{\dagger}$ follows the best-performing configuration reported in the original paper~\cite{yang2025mindjourney}, using an o-series reasoning model with SVC~\cite{zhou2025stable}, while WorldMAP uses Qwen3-VL-8B as the student backbone.}}
  \label{tab:main-2d}
  \footnotesize
  \setlength{\tabcolsep}{3pt}
  \begin{tabular}{p{0.30\columnwidth}p{0.30\columnwidth}rrr}
    \toprule
    Method & Method Family & ADE $\downarrow$ & FDE $\downarrow$ & DTW $\downarrow$ \\
    \midrule
    GPT-5.4 & \multirow{3}{*}{Proprietary} & 94.65 & 150.52 & 66.49 \\
    o3 &  & 112.14 & 177.27 & 57.30 \\
    Gemini-3-Pro &  & 51.27 & 67.19 & \textbf{31.63} \\
    \midrule
    Qwen3-VL-4B & \multirow{4}{*}{Open-source} & 140.42 & 256.00 & 136.53 \\
    Qwen3-VL-8B &  & 183.93 & 339.58 & 177.33 \\
    Qwen3-VL-32B &  & 151.21 & 298.65 & 108.06 \\
    InternVL3-14B &  & 123.33 & 218.19 & 115.01 \\
    \midrule
    MindJourney$^{\dagger}$ & \multirow{2}{*}{\shortstack[l]{World-model-\\augmented}} & 152.41 & 250.17 & 84.84 \\
    WorldMAP (Ours) &  & \textbf{42.06} & \textbf{38.87} & 31.95 \\
    \bottomrule
  \end{tabular}
\end{table}

\subsection{Analysis}
\label{sec:exp-discussion}

\noindent\textbf{Direct VLM trajectory prediction remains inconsistent.}
Strong proprietary VLMs can already produce reasonably good trajectories in 2D pixel space, but the comparison in Table~\ref{tab:main-2d} shows that direct prediction remains unreliable as a general solution. WorldMAP attains the best ADE and FDE while remaining close to the strongest proprietary baseline on normalized DTW, indicating that grounded trajectory prediction still benefits substantially from an explicit supervision pathway. The gap is even larger for open-source models, whose direct trajectory predictions degrade markedly in this setting.
\noindent\textbf{Generated views are not automatically helpful for precise trajectory prediction.}
The comparison with MindJourney is especially informative. 
In particular, MindJourney underperforms the direct o-series baseline in Table~\ref{tab:main-2d}, suggesting that additional imagined views are not automatically helpful for precise trajectory prediction. Because navigation depends on accurate geometric alignment of traversable space, target location, and stopping position, generated views that are semantically plausible yet cross-view inconsistent can introduce misleading evidence rather than useful guidance. This echoes the core finding of AVIC~\cite{yu2026and}: imagination is useful only when it is invoked appropriately, rather than applied indiscriminately.

\noindent\textbf{WorldMAP lifts a small open-source backbone into a stronger regime.}
The same open-source semantic backbone behaves very differently depending on how it is used. When asked to predict trajectories directly from egocentric observations, it performs poorly; when trained as the WorldMAP student, it becomes competitive with much stronger proprietary systems. This comparison does not isolate supervision as the only changed factor, since the teacher also contributes retrieval, grounding, and explicit planning. Nevertheless, together with the ablations below, it consistently supports the importance of the teacher-student supervision pathway in our setting.
\subsection{Ablation Studies}
\label{sec:exp-ablation}

\noindent\textbf{Effect of student backbone scale.}
We further compare WorldMAP students built on different Qwen3-VL backbones while keeping the same teacher-side pipeline and evaluation protocol fixed. This isolates how much of the final performance comes from student backbone scale itself once the model is trained under WorldMAP supervision. As shown in Table~\ref{tab:ablation-backbone}, the Qwen3-VL-8B student outperforms the 4B student on all three metrics, indicating that additional student capacity remains beneficial when the supervision pipeline is held constant. At the same time, the gain is moderate relative to the much larger improvements observed from changing the supervision recipe, suggesting that teacher-generated supervision is the dominant source of improvement in our setting.

\begin{table}[t]
  \centering
  \caption{\textbf{Ablation on WorldMAP student backbone scale.}}
  \label{tab:ablation-backbone}
  \footnotesize
  \setlength{\tabcolsep}{4pt}
  \begin{tabular}{p{0.50\columnwidth}rrr}
    \toprule
    WorldMAP student backbone & ADE $\downarrow$ & FDE $\downarrow$ & DTW $\downarrow$ \\
    \midrule
    Qwen3-VL-4B & 45.07 & 42.84 & 33.56 \\
    Qwen3-VL-8B & 42.06 & 38.87 & 31.95 \\
    \bottomrule
  \end{tabular}
\end{table}

\noindent\textbf{Effect of training supervision composition.}
We ablate the supervision used to train the student by comparing four settings: (1)~Train GT only, (2)~Train GT + WM pseudo-labels (usable), (3)~Train GT + WM pseudo-labels (usable / borderline), and (4)~WM pseudo-labels only (usable / borderline). This tests whether WM pseudo-labels are beneficial and how quality filtering interacts with Train GT. Table~\ref{tab:ablation-supervision} shows that supervision composition has a large effect: Train GT only and WM pseudo-labels only both perform poorly, while combining Train GT with teacher-generated pseudo-labels gives the strongest results. The usable-only setting yields the best ADE and FDE, while adding borderline samples slightly improves DTW.

\begin{figure}[t]
  \centering
  \includegraphics[width=\columnwidth]{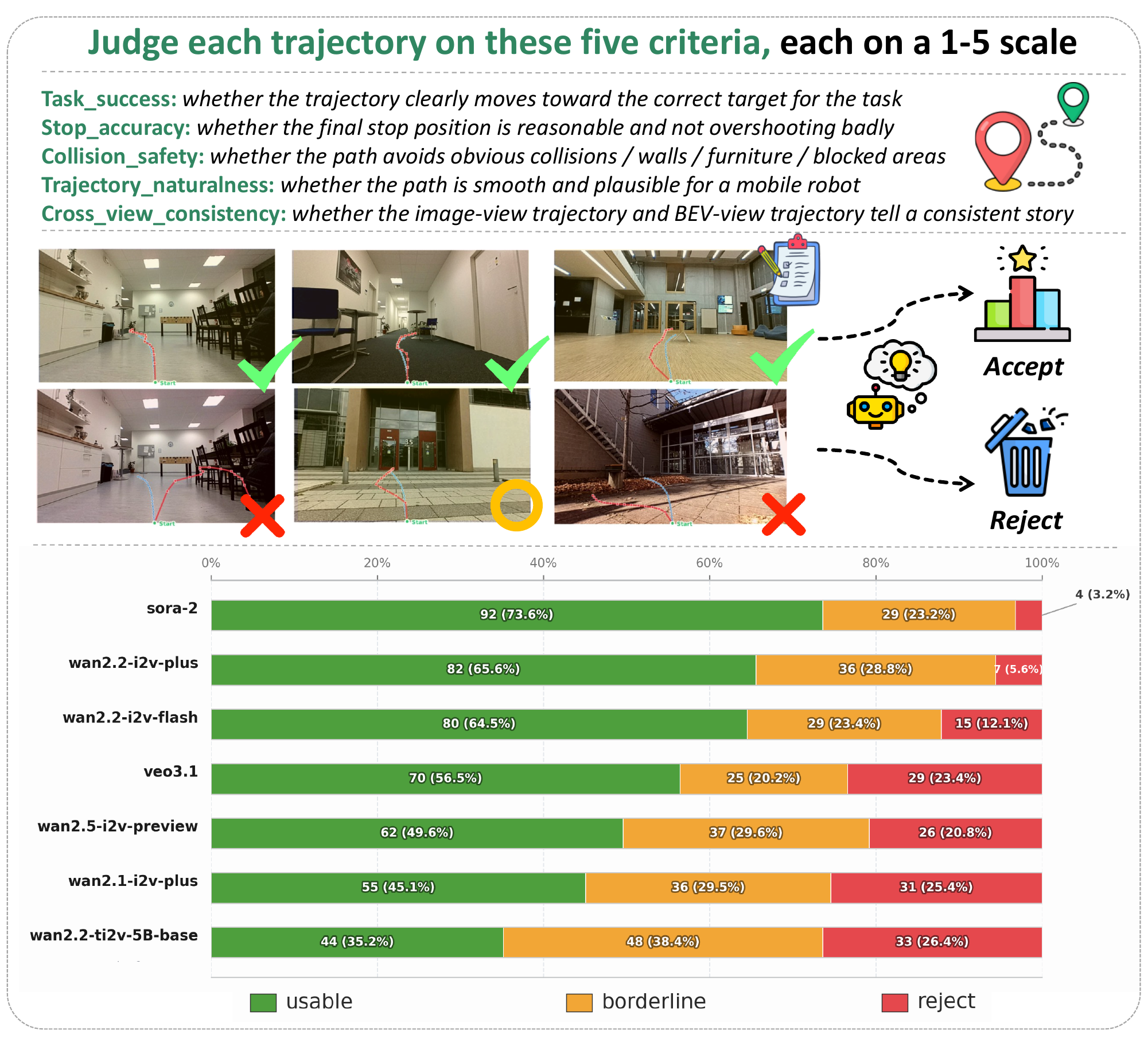}
  \caption{\textbf{VLM-based quality assessment of world-model pseudo-labels.} Generated trajectories are evaluated using five criteria: task success, stop accuracy, collision safety, trajectory naturalness, and cross-view consistency. Each trajectory is then categorized as usable, borderline, or reject, and these quality tiers are used in the supervision ablation below. The bottom panel summarizes the resulting quality distribution across world models.}
  \Description{A quality-assessment figure for world-model pseudo-labels. The top panel shows five evaluation criteria and example usable, borderline, and rejected trajectories. The bottom panel shows the proportions of usable, borderline, and rejected trajectories for different world models.}
  \label{fig:qc-judge}
\end{figure}

Figure~\ref{fig:qc-judge} summarizes the VLM-based quality-assessment protocol used to curate WM pseudo-labels. In the student training pipeline, usable trajectories form the default pseudo-label set, while borderline trajectories are treated as an optional expansion set.

\begin{table}[t]
  \centering
  \caption{\textbf{Ablation on student training supervision.}}
  \label{tab:ablation-supervision}
  \footnotesize
  \setlength{\tabcolsep}{4pt}
  \begin{tabular}{p{0.58\columnwidth}rrr}
    \toprule
    Training supervision & ADE $\downarrow$ & FDE $\downarrow$ & DTW $\downarrow$ \\
    \midrule
    WM pseudo-labels only (usable / borderline) & 95.98 & 141.10 & 88.25 \\
    Train GT only & 78.34 & 121.75 & 75.09 \\
    Train GT + WM pseudo-labels (usable / borderline) & 42.85 & 40.97 & 31.78 \\
    Train GT + WM pseudo-labels (usable) & 42.06 & 38.87 & 31.95 \\
    \bottomrule
  \end{tabular}
\end{table}

\section{Discussion}
\label{sec:discussion}

\noindent\textbf{What the current evidence supports.}
The experiments support a clear but bounded conclusion: in our setting, reliable navigation trajectory prediction depends strongly on transforming generated futures into persistent teacher representations for downstream supervision. The gains cannot be attributed to supervision alone, since WorldMAP also relies on memory retrieval, semantic grounding, and explicit planning. Rather, these components become more effective once world-model evidence is consolidated into a form aligned with navigation learning. The main contribution therefore lies not in any single module, but in the teacher--student design that converts generated futures into grounded supervision.

\noindent\textbf{A fast--slow view of the framework.}
WorldMAP can also be viewed through a fast--slow lens, related in spirit to dual-process discussions in embodied AI and recent fast--slow navigation architectures~\cite{posner2020robots,wei2025ground,zhou2025fsr}. The teacher is the slow system: it uses expensive world-model imagination, retrieval, grounding, and FMM planning to produce aligned supervision. The student is the fast system: it amortizes this process into a lightweight predictor for test-time trajectory generation. This view also helps explain our main finding: world models may be more valuable as supervision sources than as action-ready evidence, because long-horizon, cross-view navigation demands a level of consistency current world models do not yet reliably provide, and because even accurate imagined futures contain information not directly aligned with navigation. This is also reflected in the negative MindJourney result: when imagined views are not geometrically consistent enough for precise trajectory prediction, consuming them directly at test time can hurt rather than help, echoing the core finding of AVIC~\cite{yu2026and}. The key is therefore to extract supervision about targets, obstacles, and traversability rather than consume generated futures verbatim.

\noindent\textbf{Scope and limitations.}
Our results are measured on Target-Bench, which, to our knowledge, is the first benchmark specifically designed to evaluate world models for mapless navigation toward semantic targets in real-world environments~\cite{wang2025target}. It is therefore a particularly appropriate testbed for our setting and already covers diverse unstructured indoor and outdoor scenes. The gains establish effectiveness in this benchmark, but not yet transfer to highly dynamic scenes, multi-level environments, or long-horizon exploration. WorldMAP also remains constrained by the quality of future generation and geometric reconstruction: even advanced world models may produce inconsistencies, missing regions, or hallucinated content. Nevertheless, the teacher can still distill stable navigational structure through retrieval, multi-view grounding, and geometric projection, turning imperfect generations into useful supervision. Broader evaluation in more diverse environments remains necessary.

\section{Conclusion}
\label{sec:conclusion}
We presented WorldMAP, a teacher--student framework for navigation trajectory prediction that converts world-model-generated futures into persistent semantic-spatial structure and planning-derived supervision. Its world-model-driven pipeline extracts grounded trajectory pseudo-labels from generated videos and distills them into a lightweight student for direct trajectory prediction from vision-language inputs. On Target-Bench, WorldMAP achieves the best ADE and FDE among the compared methods, while lifting a small open-source VLM into a performance regime competitive with proprietary models on normalized DTW. More broadly, the results suggest that, in embodied navigation, world models may be most valuable not as sources of action-ready imagined evidence, but as engines for synthesizing aligned supervision that teaches grounded navigation behavior. This points to a promising direction for future embodied systems: using world models to construct persistent semantic-spatial structure and aligned supervision for learning, rather than relying on imagined observations as end-to-end action policies.

\bibliographystyle{IEEEtran}
\bibliography{main}

\clearpage
\phantomsection
\label{sec:supplementary}

\begin{center}
  {\Large\bfseries Supplementary Material\par}
\end{center}
\vspace{0.8em}

\appendices
\renewcommand{\thesubsection}{\thesection.\arabic{subsection}}

This supplementary material provides additional implementation details and qualitative evidence.

We focus on four aspects:
\begin{itemize}[leftmargin=1.5em]
  \item how raw Target-Bench trajectories are converted into projected 2D trajectory references;
  \item how real-frame trajectories are aligned with the world-model frame through homography;
  \item how teacher supervision is constructed from multi-view grounding and BEV planning;
  \item how the student consumes teacher supervision and how the final predictor behaves qualitatively.
\end{itemize}

\section{Data Processing Details}
\label{sec:supp-data}

This section projects Target-Bench trajectories onto the real-world first frame, aligns them with the world-model first frame, and summarizes the processed data.

\begin{figure}[!t]
  \centering
  \includegraphics[width=\columnwidth]{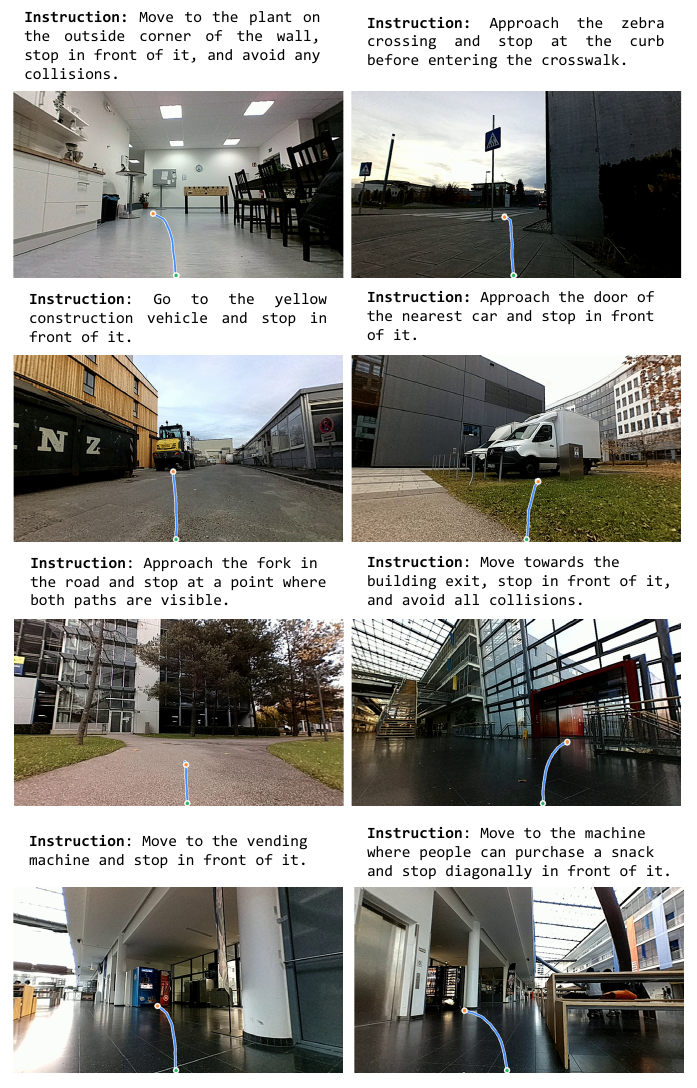}
  \caption{Representative examples of projecting Target-Bench 3D SLAM trajectories onto 2D real-world first frames.}
  \Description{A gallery of representative indoor and outdoor scenes showing instructions together with projected ground-truth trajectories overlaid on the real-world first frame.}
  \label{fig:supp-gt-gallery}
\end{figure}

Each Target-Bench sample contains a first-person observation, a language instruction, and a SLAM-estimated 3D robot trajectory.
\noindent\textbf{Trajectory Projection.}
As part of preprocessing, we project the 3D SLAM trajectory onto the real-world first frame to obtain a 2D pixel-space ground-truth trajectory.
Figure~\ref{fig:supp-gt-gallery} shows representative projected trajectories on the real-world first frames before homography-based transfer.

\begin{figure}[!t]
  \centering
  \includegraphics[width=0.94\columnwidth]{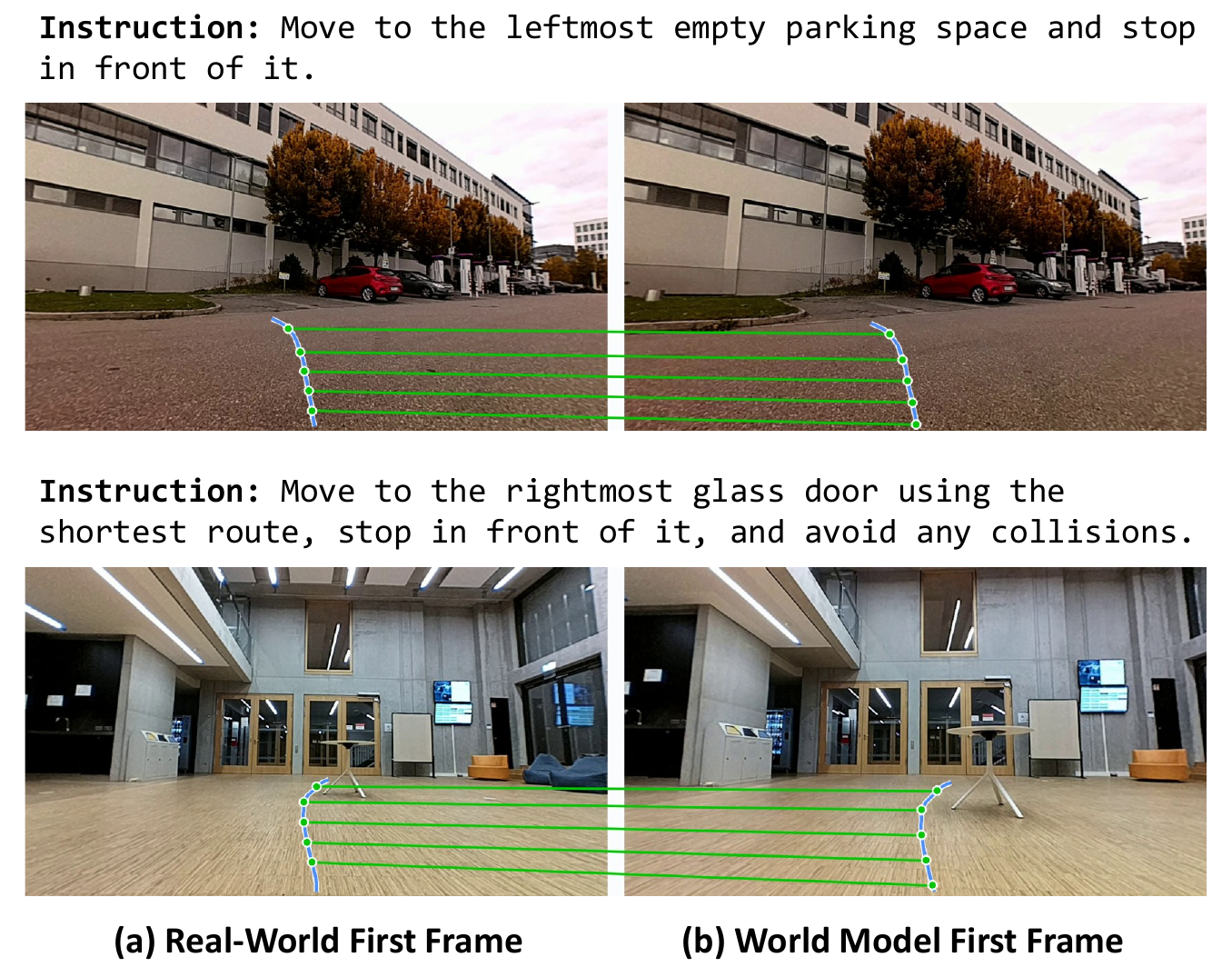}
  \caption{Illustration of our homography protocol.}
  \Description{Two example rows showing homography-based transfer between the real first frame and the world-model first frame. In each row, the left image is the real-world first frame with the projected ground-truth trajectory, and the right image is the corresponding world-model first frame with the homography-transferred trajectory. Green lines connect corresponding waypoints across the two frames.}
  \label{fig:supp-gt-homography}
\end{figure}

\noindent\textbf{Homography-Based Alignment.}
Since the real-world first frame and the world-model first frame are not directly aligned, we use a scene-specific homography protocol to establish correspondence between the two views.
For each scene, we estimate this mapping automatically with ORB feature matching and a RANSAC-based homography fit.
In practice, we keep up to 400 matched features, require at least 20 inliers, and reject unstable fits whose transferred image center shifts excessively.
The accepted homography transfers the projected trajectory from the real-world first frame to the corresponding world-model first frame and also supports inverse mapping back to the benchmark frame.
This alignment is used only during preprocessing; downstream planning remains in the world-model frame.
Figure~\ref{fig:supp-gt-homography} shows two examples of this transfer process and the corresponding waypoint matches.

\begin{figure}[!t]
  \centering
  \includegraphics[width=0.84\columnwidth]{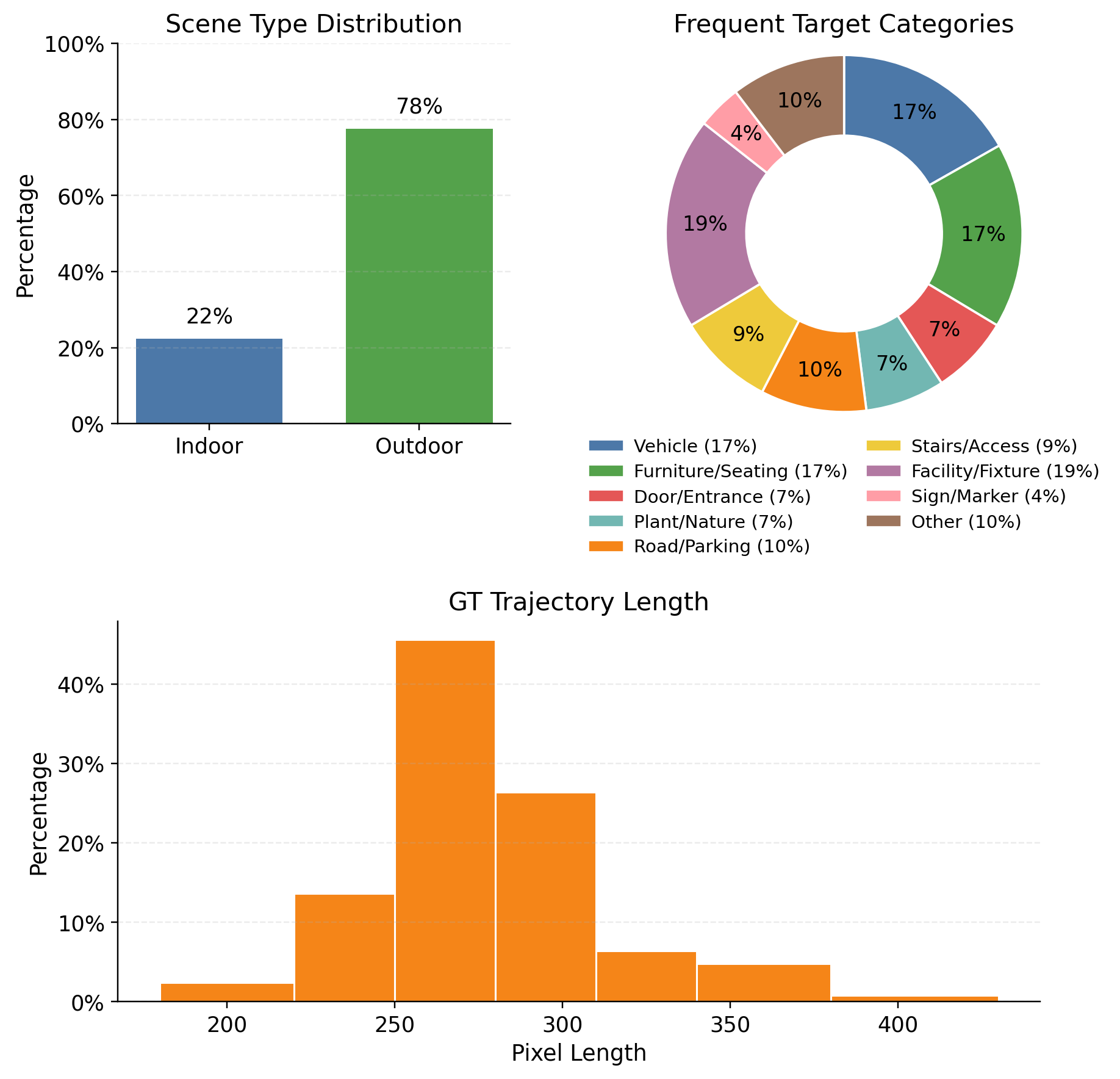}
  \caption{Statistics of the processed data.}
  \Description{Three bar-chart style plots summarizing the processed validation split. The left plot shows the counts of indoor and outdoor scenes. The middle plot shows the counts of target categories mentioned in the instructions, including vehicles, furniture or seating, doors or entrances, plants or nature, road or parking targets, stairs or access structures, facilities or fixtures, and signs or markers. The right plot is a histogram of projected ground-truth trajectory lengths in image space.}
  \label{fig:supp-dataset-stats}
\end{figure}

\noindent\textbf{Visualization and Statistics.}
Figure~\ref{fig:supp-dataset-stats} summarizes the processed validation split in terms of the indoor/outdoor ratio, target-category composition, and projected trajectory-length distribution.

\section{Implementation Details}
\label{sec:supp-impl}

\subsection{Teacher Pipeline}
\label{sec:supp-teacher}

This subsection summarizes the teacher-side details most relevant to how supervision is constructed.

\noindent\textbf{Scene Construction and Navigation Plane.}
Each world-model scene provides generated frames together with DA3-derived camera metadata.
A semantic memory built from these views supports retrieval, and a single navigation plane estimated from valid depth backprojections with a RANSAC fit defines the shared geometric reference for grounding and planning.

\begin{figure}[t]
  \centering
  \includegraphics[width=0.94\columnwidth]{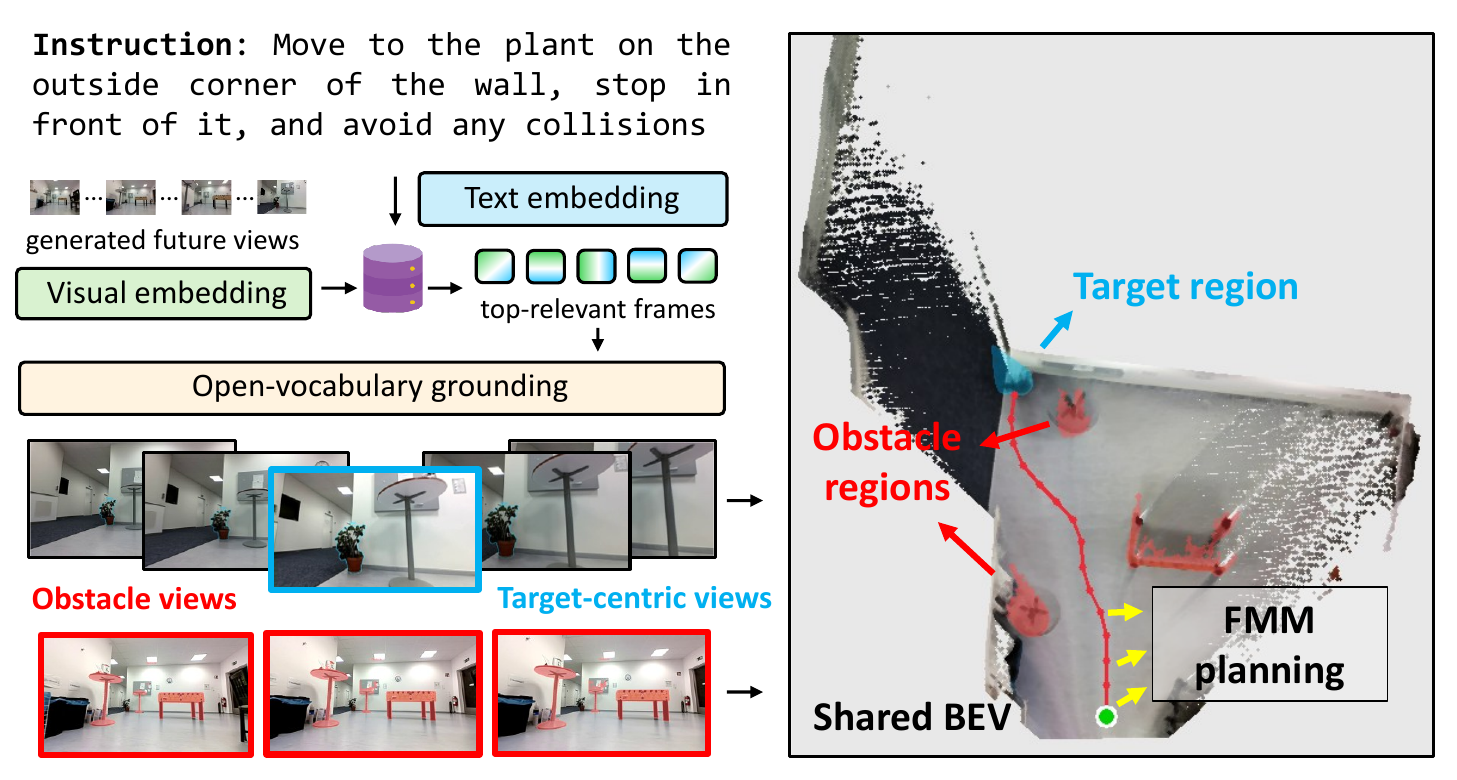}
  \caption{Intermediate stages of the teacher pipeline.}
  \Description{A summary diagram of the teacher pipeline. Generated future views are encoded into semantic memory, retrieved for target and obstacle grounding, fused into a shared bird's-eye-view representation, and used by FMM path planning to produce trajectory supervision.}
  \label{fig:supp-teacher-overview}
\end{figure}

\noindent\textbf{Task-Aware Grounding and Multi-View Fusion.}
The teacher first retrieves semantically relevant memory views and rewrites the instruction into a segmentation-ready target phrase.
Target and obstacle regions are then segmented with UniPixel-7B on the retrieved generated frames.
For target grounding, we start from the top-retrieved views, apply a VLM-based mask judge, and fuse a small set of reliable candidates.
For obstacle grounding, heavily over-segmented masks are discarded.
The selected masks are backprojected with DA3 depth and camera poses, merged in 3D, and reprojected both to the first frame and to a shared BEV map.
In BEV, target regions are consolidated into compact support areas, while obstacle regions remain spatially separated.
Figure~\ref{fig:supp-teacher-overview} summarizes this transition from memory-based retrieval and open-vocabulary grounding to BEV-based planning.

\noindent\textbf{BEV Planning and Teacher Supervision.}
After target and obstacle projection, the planner builds a BEV cost map at 5 mm per cell over a vertical band from 0.5 m below to 1.5 m above the estimated plane.
FMM planning treats projected obstacles and BEV background as blocked structure, applies a small safety margin around occupied regions, and increases traversal cost near blocked or uncertain cells.
The goal is derived from grounded target geometry by combining the nearest accessible boundary region with the target centroid.
The resulting path is smoothed, resampled, and used as teacher trajectory supervision.

\subsection{Student Distillation}
\label{sec:supp-student}

This subsection summarizes the student configuration used in our experiments.

Our student follows a two-stage setup.
Stage A adapts a Qwen3-VL-8B backbone with a LoRA adapter so that the first-person RGB observation and language instruction are encoded into geometry-aware text and vision features rather than directly decoded into waypoint text.
Stage B trains a lightweight trajectory decoder on top of these multimodal features.

The decoder combines the normalized start point with pooled text, vision, and cross-modal features and predicts multiple future waypoint hypotheses in image space.
Our trajectory decoder uses a hidden dimension of 256, three MLP layers, dropout 0.1, a 256-dimensional pooled feature projection, and 8 cross-attention heads.
Teacher-generated trajectories provide the supervision.

Training uses AdamW with learning rate $5\times 10^{-4}$, weight decay $10^{-4}$, batch size 64, and 100 epochs with a cosine scheduler.
All training and inference experiments were conducted on a machine with four NVIDIA A100 80GB GPUs.
At inference time, the student runs without the teacher and returns one final trajectory selected from the predicted hypotheses using the same best-of-$K$ formulation as in the main paper.

\section{Additional Qualitative Results}
\label{sec:supp-qualitative}

\begin{figure*}[t]
  \centering
  \includegraphics[width=0.96\textwidth]{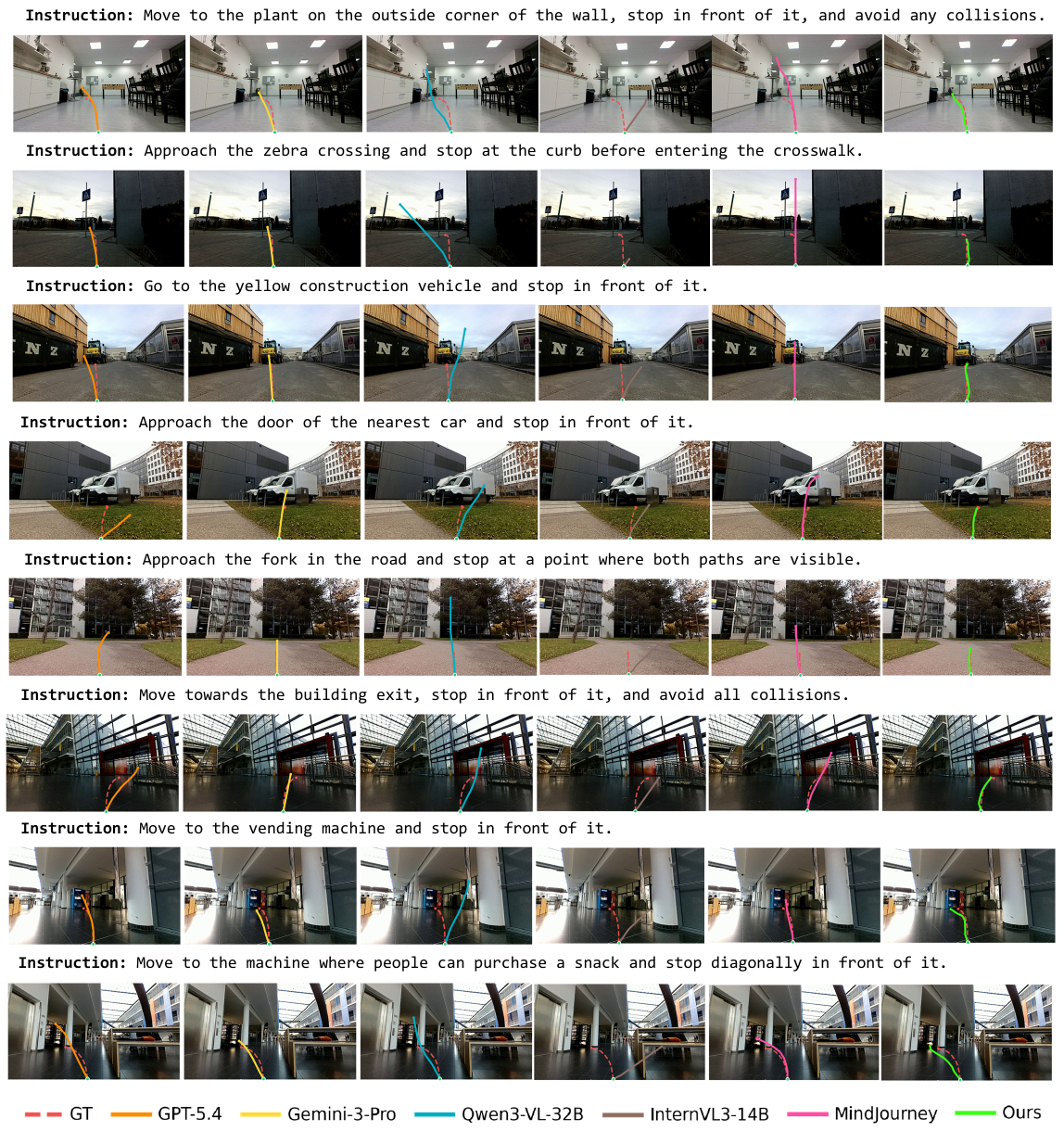}
  \caption{Additional qualitative comparisons across models.}
  \Description{A multi-row qualitative comparison figure for navigation results. Each row corresponds to one instruction and visualizes predicted trajectories from several models together with the ground-truth path. The figure shows representative cases in which the proposed method tracks the target more accurately and stops closer to the intended endpoint than competing methods.}
  \label{fig:supp-main-results}
\end{figure*}

\noindent\textbf{Qualitative Comparison Setup.}
Figure~\ref{fig:supp-main-results} shows additional qualitative results of the final predictor on representative indoor and outdoor scenes.
Each row corresponds to one instruction, and the columns compare final predicted trajectories under the same start view.

\noindent\textbf{Qualitative Observations.}
Several rows illustrate complementary strengths of our method.
In the zebra-crossing, nearest-car-door, and snack-machine cases, our predictor stops closer to the intended curb, door-facing region, or diagonal stopping pose instead of merely moving toward the correct semantic category.
In the plant, fork-in-the-road, and building-exit cases, it also approaches from a more plausible direction and follows a trajectory shape that better matches the scene layout.

\noindent\textbf{Common Failure Patterns in Baselines.}
The compared baselines show larger variance across the same cases.
Typical failure modes include drifting toward the wrong target, overshooting the destination, stopping with a large offset, and approaching from an implausible heading even when the coarse target category is correct.
As a result, some trajectories reach the right area but still fail to satisfy the instruction at the level of stopping position, approach side, or route shape.

\end{document}